\pdfoutput=1
\documentclass{article}
  \PassOptionsToPackage{numbers, compress, square, sort, comma}{natbib}

 \usepackage[preprint]{neurips_2019}

\usepackage[utf8]{inputenc} 
\usepackage[T1]{fontenc}    
\usepackage{hyperref}       
\usepackage{url}            
\usepackage{booktabs}       
\usepackage{amsfonts}       
\usepackage{nicefrac}       
\usepackage{microtype}      

\usepackage{graphicx}
\usepackage{subfigure}
\usepackage{hyperref}
\usepackage{placeins} 

\usepackage{amsmath}
\usepackage{amssymb}
\usepackage[export]{adjustbox}
\usepackage{arydshln} 
\usepackage{wrapfig}

\newcommand{\CD}{\mathcal{D}}
\newcommand{\EE}{\mathbb{E}}

\newcommand{\thetav}{\boldsymbol{\theta}}
\newcommand{\xv}{\boldsymbol{x}}

\newcommand{\epsilonv}{\boldsymbol{\epsilon}}

\title{Survival Function Matching for Calibrated Time-to-Event Predictions}

 \author{
	Paidamoyo Chapfuwa, Chenyang  Tao,	Lawrence  Carin , Ricardo  Henao\\
	Duke University\\
	\texttt{paidamoyo.chapfuwa@duke.edu} \\
}

\begin{document}

\maketitle

\begin{abstract}
Models for predicting the time of a future event are crucial for risk assessment, across a diverse range of applications.
Existing time-to-event (survival) models have focused primarily on preserving pairwise ordering of estimated event times, or relative risk. 
Model calibration is relatively under explored, despite its critical importance in
time-to-event applications. 
We present a survival function estimator for probabilistic predictions in time-to-event models, based on a neural network model for draws from the distribution of event
times, without explicit assumptions on the form of the distribution. 
This is done like in adversarial learning, but we achieve learning without a discriminator or adversarial objective.
The proposed estimator can be used in practice as a means of estimating and comparing conditional survival distributions, while accounting for the predictive uncertainty of probabilistic models.
Extensive experiments show that the proposed model outperforms existing approaches, trained both with and without adversarial learning, in terms of both calibration and concentration of time-to-event distributions.
\end{abstract}

\section{Introduction}
Time-to-event studies aim to characterize the covariate effects on the time of a future event, while capitalizing on information from censored events when performing learning.
Conventional nonparametric time-to-event (also called survival) models primarily involve methods that maximize the Concordance Index (C-Index) \citep{harrell1984regression}, a metric related to the receiver operating characteristic, that quantifies the degree to which estimated events result in pairwise orderings that are consistent with observed event times, \emph{i.e.}, the ground truth.
Consequently, any model that is able to estimate properly ordered but \emph{proportional} event times can score high in terms of C-Index.
A prominent example is the widely used Cox Proportional Hazards (CPH) model \citep{cox1992regression}.

Predicting temporally accurate event times is important in a variety of applications, \emph{e.g.}, risk profiling \citep{rajkomar2018scalable, hippisley2013predicting}, drug development \citep{fischl1987efficacy}, and prevention of online fraudulent activities \citep{zheng2018safe}.
Estimating temporally accurate event times typically involves the use of parametric Maximum Likelihood Estimation (MLE) approaches \citep{kleinbaum2010survival} or recently-developed nonparametric sampling based methods, \emph{e.g.}, via adversarial learning \citep{chapfuwa2018adversarial} or normalizing flows \citep{miscouridou2018deep}.
Further, given the critical time-sensitive nature of time-to-event modeling, it is highly desirable to design models that are not only temporally accurate but also produce \emph{population-calibrated} and \emph{uncertainty-aware} predictions.

Classical survival models include the CPH semiparametric model \citep{cox1992regression} that learns \emph{relative} risk (proportional to time-to-event) as a function of covariates, and the Accelerated Failure Time (AFT) model \citep{wei1992accelerated}, a parametric specification for temporally accurate event times that assumes covariates either accelerate or decelerate the progression of event time.
AFT often assumes log-normal distributed event times, however, other likelihood functions have been considered, \emph{e.g.}, exponential, Gamma, Weibull, \emph{etc}. \citep{bender2005generating,kleinbaum2010survival}.
These classical approaches assume a linear relationship between event times and covariates, which may be limiting for modern, large and highly heterogeneous datasets.

Time-to-event methods based on deep-learning are often direct extensions of classical models that aim to learn more flexible, non-linear mappings between event times and covariates.
CPH-based deep learning methods \citep{katzman2016deep,zhu2016deep} have demonstrated improvements in C-Index relative to classical approaches, in some settings.
Parametric extensions include the Deep Regularized Accelerated Failure Time (DRAFT) model \citep{chapfuwa2018adversarial}, Deep Survival Analysis (DSA) \cite{ranganath2016deep} and the Survival Continuous Ranked Probability Score (S-CRPS) model \cite{avati2018countdown}.
Nonparametric extensions include Deep Adversarial Time-to-Event (DATE) \cite{chapfuwa2018adversarial}, nonparametric DSA \cite{miscouridou2018deep} and Gaussian-process-based models \citep{fernandez2016gaussian, alaa2017gaussianmulti,lee2019temporal}.
As an alternative to a strict time-to-event formulation, some approaches discretize event times and specify models that predict the probability of survival at discrete intervals \citep{yu2011learning, fotso2018deep,lee2018deephit}.

Methods that produce uncertainty-aware predictions aim to estimate time-to-event {\em distributions}, rather than point estimates.
Most approaches, parametric or not, result in either a parametric time-to-event distribution, \emph{e.g.}, log-normal in AFT, DRAFT and S-CRPS and Weibull in DSA, or samples from an implicitly defined distribution, \emph{e.g.}, DATE and nonparametric DSA. 
The latter uses normalizing flows.
Importantly, uncertainty-aware predictions are only useful if the time-to-event distributions are concentrated, \emph{i.e.}, their probability masses have coverage much smaller than the observed time range.
This is key, because only in that case can uncertainty be leveraged effectively for ranking or prioritizing events/subjects.
However, only a few approaches have considered the uncertainty of the predictions when assessing performance, namely, \cite{chapfuwa2018adversarial} via distribution coverage and \cite{avati2018countdown} via coefficient-of-variation metrics.

Calibration, a descriptor of a predictive model that characterizes the statistical consistency of the predictions relative to the distribution of the observations on a population level, has been studied in forecasting \citep{degroot1983comparison}, Bayesian analysis \citep{dawid1982well}, and in machine learning, for classification \citep{guo2017calibration} and regression \citep{kuleshov2018accurate} problems.
Unfortunately, it is under-explored in time-to-event models.
Exceptions include \cite{vinzamuri2017pre,zhang2018nonparametric,bellot2018boosted,lee2019temporal} that use (time horizon) thresholded time-to-event Brier scores to asses calibration \citep{brier1950verification}, and \cite{avati2018countdown} that use calibration slope as a way to compare model performance.
Note that although Brier scores are often used to assess calibration, most commonly in classification models, summaries of calibration curves such as the calibration slope are usually considered more informative \citep{steyerberg2010assessing}.

We present an approach that \emph{implicitly} defines time-to-event distributions conditioned on covariates via a neural network specification, from which we can synthesize temporally accurate, concentrated and calibrated time-to-event distributions.
To this end, $i$) we present a reinterpretation of the Kaplan-Meier estimator for survival functions; $ii$) we extend it to estimate survival functions \emph{conditional} on covariates; $iii$) we show that the new estimator can be used for visual and quantitative assessment of calibration; $iv$) we propose using it as an objective function in a neural-network-based nonparametric time-to-event model, to encourage calibrated predictions; $v$) we directly match the conditional survival function of the model to that of the ground truth without the need of adversarial techniques \citep{goodfellow2014generative}; $vi$) we show that our \emph{survival function matching} approach is related to earth mover's distance minimization; and $vii$) we present extensive quantitative and qualitative results, showing that our approach outperforms existing time-to-event models in terms of calibration, while being competitive in terms of C-Index and concentration (sharpness) of the predicted time-to-event distributions.

\section{Background}
Assume a time-to-event dataset, $\CD=\{\xv_n,t_n,y_n\}_{n=1}^N$, consisting of $N$ observations (or subjects).
For the $n$-th observation, we have $d$ covariates, $\xv_n = [x_{1n}, \ \ldots, \ x_{dn}] \in \mathbb{R}^{d}$, a time point, $t_{n}$, and a censoring indicator, $y_{n} \in \{0, 1\}$.
When $y_{n}=1$, $t_{n}$ represents the \emph{time-to-event} of interest, and when $y_{n}=0$, $t_{n}$ is the censoring time.
Typically, events are \emph{right} censored, meaning that given $y_{n}=0$, all we know about the $n$-th observation is that we have not observed the event of interest up to time $t_{n}$.
Though \emph{left} and \emph{interval} censored events are possible, these are far less common and are thus not usually considered in practice.
Here we only consider right censoring, however, the proposed approach is general and can be readily extended using ideas from \citep{avati2018countdown}.

Time-to-event (or survival) models either characterize the conditional survival function $S(t|\xv)$, time density $f(t|\xv)$, or the hazards function $h(t|\xv)$, where the conditioning is on covariates $\xv$.
The survival function $S(t|\xv) = P(\tau > t| \xv)$, for $\tau>0$, which can also be written as the complement of the conditional cumulative density function, $F(t|\xv)$; hence, $S(t|\xv)=1-F(t|\xv)$ is a monotonically decreasing function of time.
Learning the time-to-event conditional distribution, $f(t|\xv)$, can in principle yield both $S(t|\xv)$ and $h(t|\xv)$, provided $f(t|\xv)=S(t|\xv)h(t|\xv)$.
%
%
For some parametric choices of the conditional density, $f(t|\xv)$, the survival and hazards functions can be obtained in closed-form \citep{kleinbaum2010survival}.
For instance, assuming the exponential density, $f(t| \xv)  = \lambda_{\xv} \exp ( - \lambda_{\xv} t)$, yields $h(t| \xv) =\lambda_{\xv}$ and $S(t| \xv) = \exp ( - \lambda_{\xv} t)$, where $\lambda_x$ is a function of $\xv$.
See \citep{bender2005generating} for a few other examples.
In practice, we seek to approximate the time density $f(t|\xv)$ with $q(t|\xv)$, a function parametrically or nonparametrically specified and learned from data, ${\cal D}$.
The dataset ${\cal D}$ represents the \emph{ground truth} or, conceptually, the empirical joint distribution $p(t,y,\xv)$ with marginals $p(t)$, $p(y)$ and $p(\xv)$, from which $p(t)$ is of most interest in our case, as described below.

\paragraph{The Kaplan-Meier Estimator}
The standard Kaplan-Meier (KM) estimator \citep{kaplan1958nonparametric} is a widely-used frequentist approach to estimate the (marginal) survival function, $S(t)$, using samples from $p(t)$, \emph{i.e.}, the time-to-event empirical distribution.
Let $\mathcal{T} = \{ t_i |  t_i > t_{i-1} > \ldots > t_0\}$ be the set of distinct and ordered observed event times (censored and non-censored).
The KM estimate for time $t_i$ can be evaluated recursively as
\begin{align}\label{eq:km_est}
\hat{S}_{\rm KM}(t_i) =  \left(1- \frac{d_i}{n_i}\right) \hat{S}_{\rm KM}(t_{i-1})\,,
\end{align}
where $n_i$ is the number of subjects \emph{at risk} at the beginning of follow-up interval $[t_i, t_{i+1})$, $d_i$ is the number of non-censored events that occur within the same interval, $[t_i, t_{i+1})$, and $\hat{S}_{\rm KM}(t_0) = 1$, indicating that at $t_0$ there are no observed events so $d_n=0$ and $n_0=N$.
It has been shown \cite{hosmer2011applied} that the KM estimator can be interpreted as a random process, where the number of events, $d_i$, within each discrete interval $[t_i, t_{i+1})$ can be modeled as a draw from a Binomial distribution $d_i \sim \text{Binomial}(n_i, \pi)$, with mean event rate $\pi$.
Moreover, it has been proven that KM is a consistent estimator \citep{peterson1977expressing}, \emph{i.e.}, $\sqrt{N} (\hat{S}_{\rm KM}(t) - S(t))$ converges to a Gaussian process \citep{breslow1974large}, with zero mean and covariance function recursively approximated by Greenwood's formula \citep{greenwood1926report}.
%

\paragraph{Distribution-Based Kaplan-Meier Estimator}
The standard KM estimator is a \emph{population} statistic that approximates the marginal survival distribution $S(t)$.
Consequently, KM does not explicitly accommodate the use of predictions, \emph{i.e.}, individualized (subject-level) conditional survival functions.
Considering that time-to-event methods are primarily tasked with individualized predictions of conditional time densities, $f(t|\xv)$, which can be then used to obtain conditional survival functions $S(t|\xv)$, below we present a modified KM estimator that accounts for individualized time-to-event predictions.

We first consider a KM estimator for \emph{point estimates} of $S(t|\xv)$, directly formulated from the standard KM in \eqref{eq:km_est}.
It is then extended to probabilistic, \emph{distribution} estimates of $S(t|\xv)$.
The point-estimate-based KM, denoted PKM, estimates the population survival function accounting for covariates using predictions $\hat{T}_n \sim g(\xv_n)$, where $g(\xv_n)$ is some predictive function, or a summary from a probabilistic estimate of the conditional density $f(t|\xv_n)$, \emph{e.g.}, $\hat{T}_n \sim g(q(t|\xv_n))$ where $g(\cdot)={\rm mean(\cdot)}$ and $q(t|\xv_n)$ is the approximated conditional learned from dataset ${\cal D}$.
We then write
\begin{align}\label{eq:km_ref_est}
\hat{S}_{\rm PKM}(t_i ) & = \left(1- \frac{\sum_{n:y_n=1} \mathbb{I}(t_{i-1} \le \hat{T}_n <  t_{i}) }{ N - \sum_{n=1}^{N}  \mathbb{I}(\hat{T}_n < t_{i-1}) } \right) \hat{S}_{\rm PKM}(t_{i-1})\,,
\end{align}
where $\hat{S}_{\rm PKM}(t_0) = 1$, $\mathbb{I}(a)$ is an indicator function such that $\mathbb{I}(a)=1$ if $a$ holds or $\mathbb{I}(a)=0$ otherwise.
It follows from~\eqref{eq:km_ref_est} that $\hat{S}_{\rm PKM}(t_i) = \hat{S}_{\rm KM}(t_i)$, when $\hat{T}_n$ represents an observed (ground truth) time-to-event from $p(t)$.

To account for predictive uncertainly, \emph{i.e.}, for probabilistic estimates $q(t|\xv_n)$, we extend \eqref{eq:km_ref_est} to distribution-based Kaplan-Meier (DKM) estimator.
Specifically, we write
%
\begin{align}\label{eq:km_dist_est}
\hat{S}_{\rm DKM} (t_i ) = \left( 1- \frac{\sum_{n:y_n=1} F_n(t_i|\xv_n) - F_n(t_{i-1}|\xv_n)} { N - \sum_{j=1} F_n(t_{i-1}|\xv_n) } \right) \hat{S}_{\rm DKM} (t_{t-1}) \,,
\end{align}
where $F_n(t_i|\xv_n)$ is the estimated cumulative density function for subject $n$ conditioned on covariates $\xv_n$ and evaluated at $t_i$.
Note that $\hat{S}_{\rm DKM} (t_i ) = \mathbb{E}_{q(t | \xv_1)\ldots q(t | \xv_N)}[\hat{S}_{\rm PKM}(t_i) ]$ so \eqref{eq:km_dist_est} averages over (samples of) $q(t | \xv_n)$ rather than being evaluated on summaries ({\emph e.g.}, averages) of $q(t | \xv_n)$ as in \eqref{eq:km_ref_est}.
For probabilistic estimates $q(t|\xv_n)$ of $f(t|\xv_n)$, the estimator in~\eqref{eq:km_dist_est} is attractive because it accounts for the predictive uncertainty of the model, thus on a population level, it comprehensively captures the uncertainty of the estimated conditional survival distribution.

\begin{wrapfigure}[12]{R}{0.37\textwidth}
	\begin{minipage}[T]{0.36\textwidth}
		\vspace{-4mm}
		\centering
		\includegraphics[width=\columnwidth]{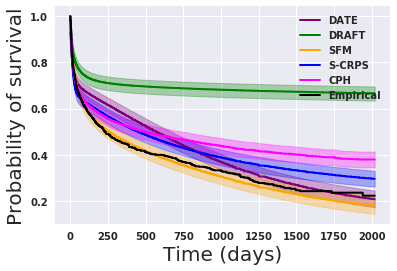}
		\vspace{-6mm}
		\caption{\small Survival function estimates for {\sc support} data. Ground truth (Empirical) is compared to test set predictions from five models.}
		\label{fg:calib_ex}
	\end{minipage}
\end{wrapfigure}

\paragraph{Calibration in Time-to-Event Models}
In the context of time-to-event modeling, calibration refers to the concept of obtaining a predictor of time-to-event (that may or may not be probabilistic) whose predictions match, on a population level, the survival distribution $S(t)$.
Figure~\ref{fg:calib_ex} shows estimated survival distributions on the {\sc support} dataset (see Section~\ref{sc:exp} for details) for five different models (DATE, DRAFT, SFM, CPH and S-CRPS) using DKM in \eqref{eq:km_dist_est}, as well as the ground truth (Empirical) using KM in~\eqref{eq:km_est}.
Error bars (shaded regions) are calculated using the exponential Greenwood's formula \citep{hosmer2011applied}.

From Figure~\ref{fg:calib_ex}, we see that DKM in \eqref{eq:km_dist_est} can be used to visually assess the calibration of estimated event times from different models relative to the ground truth.
Specifically, we see that one of the models, SFM (the proposed model) matches the ground truth (Empirical) substantially better than the alternatives (see Section~\ref{sc:exp} for details).
Strikingly, the other three models underestimate survival almost everywhere.
In the experiments, we will use KM and DKM to more directly visualize calibration, and summarize it in terms of \emph{calibration slope}.
Further, below we leverage DKM to encourage calibration during model training, \emph{i.e.}, that DKM for a given model that approximates $q(t,\xv_n)$ matches as well as possible the true survival distribution estimated via KM.
%

\section{Survival Function Matching}\label{sc:sfm}
We propose a nonparametric model for survival-function matching.
Specifically, we approximate the density $f(t|\xv)$ implicitly as $q(t|\xv)$ via deterministic function $G_{\thetav}(\xv,\epsilonv)$ which we specify as a neural network parameterized by $\thetav$ and where $\epsilonv$ is a source of stochasticity, distributed according to some simple distribution, \emph{e.g.}, uniform or Gaussian.
In this manner, we do not impose/assume an explicit form on $q(t|\xv)$, we only seek to efficiently synthesize samples from it. This type of model has been considered recently within an adversarial-learning setup \citep{chapfuwa2018adversarial}, but in the proposed work adversarial learning is not required, thus simplifying learning.
Further, \cite{chapfuwa2018adversarial} did not consider calibration.

\paragraph{Calibration objective}
Assume as above that $\mathcal{T}$ is the set of distinct and ordered observed event times (censored or non-censored).
To estimate the parameters of the model $G_{\thetav}(\xv,\epsilonv)$ that generates time-to-event samples on a population level, we match synthesized samples to the empirical survival function, $S(t)$, thus producing calibrated predictions.
We propose optimizing the following objective
%
\begin{align}\label{eq:km_emd}
\ell_{\rm cal}(\thetav;\CD) = \frac{1}{|\mathcal{T}|} \sum_{t_i \in \mathcal{T} } \left\| \hat{S}_{\rm PKM}^{p(t)}(t_i) - 	\hat{S}_{\rm PKM}^{G_{\thetav}(\xv,\epsilonv)}(t_i) \right\|_1 \,,
\end{align}
where $|{\cal T}|$ is the cardinality of ${\cal T}$, and $\hat{S}_{\rm PKM}^{p(t)}(t_i)$ and $\hat{S}_{\rm PKM}^{G_{\thetav}(\xv,\epsilonv)}(t_i)$ are obtained from $p(t)$ and samples from $G_{\thetav}(\xv,\epsilonv)$, respectively.
This is connected to KM, because $\hat{S}_{\rm PKM}^{p(t)}(t_i)$ are obtained from $p(t)$.

The objective in~\eqref{eq:km_emd} seeks to obtain model parameters, $\thetav$, for which model and empirical survival functions match.
Note that the objective accounts for both censoring and non-censored events.
Provided that the conditional survival distribution $S(t|\xv) = P(\tau > t| \xv) $ for $\tau \ge 0$ is the complement of conditional cumulative density function $F(t|\xv)$, matching the conditional survival function also matches the conditional time density $f(t|\xv)$, \emph{i.e.}, the time-to-event distribution.
%

Learning with~\eqref{eq:km_emd} is challenging because $\ell_{\rm cal}(\thetav;\CD)$ is a discrete function, and thus backpropagation is difficult.
Several techniques have been developed to efficiently obtain unbiased and low-variance gradients for backpropagation with discrete objectives or sampling distributions, thus alleviating some of its challenges.
Such techniques include REINFORCE \citep{williams1992simple}, reparameterization tricks \citep{rezende2014stochastic, kingma2013auto}, and more recently RELAX \citep{grathwohl2017backpropagation}, a technique that combines both REINFORCE and reparameterization tricks via a variance-reduction neural network.

To circumvent the challenges of optimizing over the discrete function in \eqref{eq:km_ref_est} and favor simplicity, we instead optimize over its expectation in \eqref{eq:km_dist_est}, which is continuous.
However, replacing \eqref{eq:km_ref_est} with \eqref{eq:km_dist_est} is not only inefficient, as it requires generating multiple samples from $G_{\thetav}(\xv,\epsilonv)$, but also challenging because $F(t|\xv)$, the conditional cumulative function for $G_{\thetav}(\xv,\epsilonv)$ is not available in closed-form.
Conveniently, we can replace the indicator functions $\mathbb{I}(a)$ in \eqref{eq:km_ref_est} with Heaviside step functions, $H(b) = \tfrac{1}{2}(\text{sign}(b)+1)$, therefore obtaining a differentiable formulation:
\begin{align}\label{eq:km_ref_est_step}
\hat{S}_{\rm PKM}(t_i) = \left(1- \frac{\sum_{n:y_n=1} H(\hat{T}_n - t_{i-1}) - H(\hat{T}_n  - t_i)}{ N  - \sum_{n=1}^{N}  H(t_{i-1} -\hat{T}_n ) } \right) \hat{S}_{\rm PKM}(t_{i-1}) \,.
\end{align}
When evaluating the objective, $\ell_{\rm cal}(\thetav;\CD)$ in~\eqref{eq:km_emd}, $\hat{T}_n$ is either a sample from the model, $\hat{T}_n=G_{\thetav}(\xv_n,\epsilonv)$, or an observed time $\hat{T}_n\sim p(t)$, for $\hat{S}_{\rm PKM}^{G_{\thetav}(\xv,\epsilonv)}(t_i)$ or $\hat{S}_{\rm PKM}^{p(t)}(t_i)$, respectively.

\paragraph{Accuracy objective}
The objective $\ell_{\rm cal}(\thetav;\CD)$ in~\eqref{eq:km_emd} optimizes over a population estimate that encourages calibration.
However, calibration alone does not result in time-to-event samples from $G_{\thetav}(\xv,\epsilonv)$ that are accurate or concentrated wrt the ground truth.
This happens because, for a given problem, there exist many solutions that yield well-calibrated predictions that are not necessarily accurate, thus not practically useful.
For instance, take the extreme case for which a model learns to estimate $p(t)$ independent of (ignoring) the covariates, $\xv$, thus effectively recovering the KM estimator in~\ref{eq:km_est}.
So motivated, we also specify accuracy-enforcing objective functions for censored and non-censored observations by borrowing from the recently proposed DATE model \citep{chapfuwa2018adversarial}.
Specifically, we split the dataset $\mathcal{D}$ into two disjoint sets $\mathcal{D}_{c}$ and $\mathcal{D}_{nc}$, for censored and non-censored observations, respectively, and let $(t, \xv) \sim p_c$ and $(t, \xv) \sim p_{nc}$ represent, respectively, empirical distributions for these sets.
We write objective functions for $\mathcal{D}_{c}$ and $\mathcal{D}_{nc}$ as
%
%
%
\begin{align}\label{eq:date}
\ell_{\rm acc}(\thetav;\mathcal{D}_{c},\mathcal{D}_{nc}) = \mathbb{E}_{(t,\xv)\sim p_{c},\epsilonv\sim p_\epsilon}[\max(0, t-G_{\thetav}(\xv,\epsilonv)] + \EE_{(t, \xv) \sim p_{nc},\epsilonv\sim p_\epsilon} [|t - G_{\thetav}(\xv,\epsilonv)|] \,,
\end{align}
where $\epsilonv\sim p_\epsilon$ has a simple distribution (uniform or Gaussian), $\max(0,\cdot)$ in the first term encourages that time-to-event samples from the model, evaluated on censored observations $y_n=0$, are larger than the censoring time.
The second term, absolute error, encourages time-to-event samples to be accurate, \emph{i.e.}, as close as possible to the ground truth, for non-censored (observed) observations.

\paragraph{Consolidated objective}
The complete objective function for the proposed Survival Function Matching (SFM) model is $\ell(\thetav;\mathcal{D}) = \ \ell_{\rm cal}(\thetav;\mathcal{D})+ \lambda\ell_{\rm acc}(\thetav;\mathcal{D}_{c},\mathcal{D}_{nc})$,
%
%
where $\lambda > 0$ is a free parameter controlling the trade-off between the accuracy objective and the survival function matching objective in \eqref{eq:km_emd}.
In the experiments we let $\lambda=1$, however, $\lambda$ can be optimized by grid search if desired.

The complete objective is optimized using stochastic gradient descent on minibatches from $\mathcal{D}$.
Note that $\ell_{\rm cal}(\thetav;\CD)$ is a population-level objective that may be affected by the minibatch size, however, empirically we did not observe substantial differences in the performance metrics when varying the minibatch size (see the Supplementary Material).
We justify the model being insensitive to the minibatch size owing to the insight that learning with minibatches can be understood as encouraging the model to be calibrated for every minibatch, thus consequently also encouraging global calibration.

\section{Related Work}
%
Existing calibration literature in predictive models has primarily focused on recalibration techniques for predictions from classification \citep{guo2017calibration} or regression models \citep{kuleshov2018accurate}.
For classification tasks, the Brier score \citep{brier1950verification} is a commonly used proper score metric, quantifying 
the accuracy of probabilistic predictions, and thus it is often used to assess calibration.
The Brier score has also been used to asses calibration in time-to-event models \cite{vinzamuri2017pre,bellot2018boosted,zhang2018nonparametric,lee2019temporal}, however, this score has to be evaluated at pre-specified (thresholded) time horizons.
Alternatively, S-CRPS \cite{avati2018countdown} considers the integral of the Brier score evaluated at all possible thresholds \cite{gneiting2007strictly}, which is a more principled and comprehensive approach than calibration at pre-specifying time horizon thresholds.

The approach presented here is inspired by \cite{avati2018countdown}.
They considered calibration slope as a metric for evaluating performance in time-to-event models.
However, our formulation is very different from that of \cite{avati2018countdown}, in the sense that they encourage calibration by optimizing a proper score rule, the Continuous Ranked Probability Score (CRPS), whereas we tackle it directly as a survival-distribution-matching problem.
In the experiments in Section~\ref{sc:exp}, we show empirically that our more direct approach to calibration consistently outperforms CRPS.

Our work is related to other recently proposed time-to-event approaches, including Survival CRPS (S-CRPS) \citep{avati2018countdown}, that uses an AFT (log-normal distribution) specification; the \emph{conditional-GAN} approach \citep[DATE,][]{chapfuwa2018adversarial}; the AFT-based DRAFT model \citep{chapfuwa2018adversarial}; the Weibull-based Deep Survival Analysis (DSA) \citep{ranganath2016deep}; nonparametric DSA based on \emph{normalizing flows} \citep{miscouridou2018deep}; and Gaussian-process-based models \citep{fernandez2016gaussian,alaa2017gaussianmulti,lee2019temporal}.
Interestingly, excluding approaches that address \emph{thresholded} calibration with Brier-scores \cite{vinzamuri2017pre,bellot2018boosted,zhang2018nonparametric,lee2019temporal}, only S-CRPS \citep{avati2018countdown} considers \emph{global} calibration as a performance metric.
All the others focus on accuracy-centric performance estimates, \emph{e.g.}, C-Index and relative absolute error.
%

%
\emph{Optimal mass transport} approaches for distribution matching in machine learning tasks have received considerable attention recently \citep{chen2018improving, salimans2018improving}.
For one-dimensional problems, it has been shown that the characterization of the $p$-Wasserstein metric has a simple form \citep{kolouri2017optimal} $W_p(P, Q) = ( \int_{0}^{1} |F^{-1}(z) - G^{-1}(z)|^p dz)^{1/p}$
%
%
where, $F(z)^{-1}$ and $G(z)^{-1}$, for $z\in(0,1)$, are the \emph{quantile functions} of $p(t)$ and $q(t)$, respectively, and $F(t)$ and $G(t)$, their corresponding cumulative density functions.
Interestingly for $p=1$, $W_p(P, Q)$ is also known as the \emph{Monge-Rubenstein metric} \citep{villani2008optimal} or the \emph{earth mover's distance} \citep{rubner2000earth}, and it is essentially the absolute difference between the quantile functions for $p(t)$ and $q(t)$.
By contrast, the SFM objective in~\eqref{eq:km_emd} is the absolute difference between the cumulative density functions for $p(t)$ and $q(t)$, provided that $F(t)=1-S(t)$.
As a result, minimizing \eqref{eq:km_emd} and $W_p(P, Q)$ are closely related approaches to matching $p(t)$ and $q(t)$.
However, the latter explicitly imposes survival-distribution matching, which we consider more appropriate considering the goal is to obtain calibrated predictions in the context of time-to-event modeling.

\section{Experiments}\label{sc:exp}
We qualitatively and quantitatively compare the proposed approach, SFM, against DATE \citep{chapfuwa2018adversarial}, DRAFT \citep{chapfuwa2018adversarial}, and CPH \citep{cox1992regression} and S-CRPS \citep{avati2018countdown}.
Complete details of model architectures, optimization, validation and testing are in the Supplementary Material.

\paragraph{Datasets}
\begin{wraptable}[9]{T}{0.59\textwidth}
	\vspace{-6mm}
	\caption{Summary statistics of the datasets used in the experiments. The time range, $t_{\rm max}$, is noted in days except for {\sc seer} for which time is measured in months.}
	\label{tb:data}
	\vspace{1mm}
	\centering
	\begin{scriptsize}
		\begin{tabular}{lrrrrr}
			& {\sc ehr} & {\sc flchain} & {\sc support} & {\sc seer}  & {\sc sleep}\\
			\toprule
			Events (\%) & 23.9 & 27.5 & 68.1 & 51.0  & 23.8 \\
			$N$ & 394,823 & 7,894 & 9,105 & 68,082 & 5026 \\
			$d$ (cat) & 729 (106) & 26 (21) & 59 (31) & 789 (771) & 206 \\
			Missing (\%) & 1.9 & 2.1 & 12.6 & 23.4  & 18.2 \\
			$t_{\rm max}$ & 365 & 5,215 & 2,029 & 120 & 5,794 \\
			\bottomrule
		\end{tabular}
	\end{scriptsize}
\end{wraptable}
We consider five diverse datasets:
$i$) {\sc flchain}: a public dataset investigating non-clonal serum immunoglobin free light chains effects on survival time \citep{dispenzieri2012use}.
$ii$) {\sc support}: a public dataset for a survival-time study of seriously-ill hospitalized adults \citep{knaus1995support}.
$iii$) {\sc seer}: a public dataset provided by the Surveillance, Epidemiology, and End Results (SEER) Program.
We restrict the dataset to a 10-year follow-up breast cancer subcohort with three competing risks (breast cancer, cardiovascular and others). See \cite{ries2007cancer} for preprocessing details.
$iv$) {\sc ehr}: a large study from the Duke University Health System centered around multiple inpatient visits due to comorbidities in patients with Type-2 diabetes \citep{chapfuwa2018adversarial}.
$v$) {\sc sleep}: a subset of the Sleep Heart Health Study (SHHS) \citep{quan1997sleep}, a multi-center cohort study implemented by the National Heart Lung \& Blood Institute to determine the cardiovascular and other consequences of sleep-disordered breathing.

Table~\ref{tb:data} presents summary statistics of the datasets, where $d$ denotes the size of the \emph{individual} covariate vector $\xv$ after one-hot encoding for categorical (cat) variables.
\emph{Events} indicates the proportion of the non-censored events, \emph{i.e.}, the events of interest for which $y_i = 1$.
Missing indicates the proportion of missing entries in the $N \times d$ covariate matrix, and $t_{\rm max}$ is the time range for both censored and non-censored events.
For all datasets except {\sc seer}, that uses months, events are measured in days.
In the experiments we do not convert time to a common scale and model it \emph{as is}.

Details of the public datasets: {\sc flchain}, {\sc support} and {\sc seer}, including preprocessing procedures, are provided in the above references.
The other two datasets, {\sc ehr} and {\sc sleep} are not public but can be obtained upon request, see \citep{chapfuwa2018adversarial} and \citep{zhang2018national}, respectively.
For {\sc sleep} we focus on the baseline clinical visit and aggregated demographics, medications and questionnaire data as covariates.

As shown in Table \ref{tb:data}, survival datasets often contain substantial missingness, \emph{e.g.} up to 23\% in SEER data.
Interestingly, \citep{miscouridou2018deep} showed via the information-theoretic data processing inequality that there is no additional information to be gained by actively imputing missing values during training with an autoencoding arm, when compared to a simpler \emph{pre-imputation} approach in which missing values are imputed with median and mode for continuous and categorical covariates, respectively.
In view of this, here we adopt a pre-imputation strategy.
%

\vspace{-3mm}
\paragraph{Quantitative evaluation}
\begin{wraptable}{c}{0.6\textwidth} \label{table:quant_results}
	\vspace{-2mm}
	\centering
	\caption{\small Performance metrics. SFM is the proposed model.}
	\label{tb:quant}
	\vspace{1mm}
	\begin{scriptsize}
		\begin{tabular}{lrrrrr}
			& {\sc ehr} & {\sc flchain} & {\sc support} & {\sc seer}  & {\sc sleep}\\
			\toprule
			Calibration slope & & & & & \\
			\hdashline
			DATE  &0.7537  &0.9668 &0.9068 & 0.9161 & 0.9454 \\
			DRAFT  &3.2138  & 5.4183 & 2.9640 & 2.0763 &  25.2855\\
			S-CRPS &1.6246 & 1.9662& 1.1795&1.1613 & 2.5746\\
			CPH &2.5543&1.9116&1.3909&1.4358& 3.8278\\
			SFM   &   \textbf{0.7734} & \textbf{0.9807}  &   \textbf{0.9405} & \textbf{0.9540} & \textbf{1.0235} \\
			\toprule
			Mean CoV & & & & &\\
			\hdashline
			DATE  & \textbf{0.2477} & \textbf{0.3585} & \textbf{0.2987}& \textbf{0.1485} & 0.5168\\
			DRAFT  &  5.0305 &   6.2952  & 3.8689 &  3.4501 & 8.4918 \\
			S-CRPS &0.8585&0.9412& 0.7351& 0.6036 & 1.0240 \\
			CPH &-&-&-&-&-\\
			SFM   & 0.2953 & 0.4484 &  0.3930 &  0.1993 &  \textbf{0.5045} \\
			\toprule
			C-Index  & & & & &\\
			\hdashline
			DATE  & 0.7756 & 0.8264&0.8421 &  \textbf{0.8320} & 0.7416 \\
			DRAFT  &  \textbf{0.7796} &0.8341 & 0.8560 & 0.8310 & \textbf{0.7617}\\
			S-CRPS & 0.7704& 0.8286& \textbf{0.8685} & 0.8298 & 0.7529 \\
			CPH &0.7542& \textbf{0.8344}&0.8389&0.8223&0.6435\\
			SFM   &  0.7786& 0.8318 &  0.8319 &  0.8314  &  0.7491\\
			\bottomrule
		\end{tabular}
	\end{scriptsize}
	\vspace{-3mm}
\end{wraptable}
For a comprehensive quantitative evaluation of time-to-event models we consider three metrics that highlight different aspects of model performance: $i$) Concordance Index (C-Index) \citep{harrell1984regression} to quantify preservation of pairwise orderings wrt ground truth events,
$ii$) Coefficient of Variation (CoV) to assess uncertainty concentration by quantifying the dispersion of estimated time-to-event distributions, and 
$iii$) Calibration to asses the statistical consistency of the conditional survival distribution learned by a model relative to that of the ground truth.
As previously discussed, a high-performing model is one that not only preserves pairwise ordering of event times, but also results in concentrated and well-calibrated time-to-event distributions.
As discussed below, SFM outperforms other approaches in terms of calibration while being competitive in terms of C-Index (time ordering) and CoV (concentration).

{\it Calibration}:
We evaluate calibration both visually and quantitatively.
For the visual assessment, we plot the conditional survival distributions estimated from the model predictions using DKM in~\eqref{eq:km_dist_est} and compare it with the empirical survival distribution (ground truth) using KM in~\eqref{eq:km_est}, as shown in Figure~\ref{fg:calib_ex}.
Alternatively, we plot the estimated conditional cumulative density function for each model using $1 - \hat{S}_{DKM}(t_i)$ against the marginal cumulative density function for the ground truth using $1 - \hat{S}_{KM}(t_i)$.
In both cases, $t_i\in{\cal T}$.
If the estimated cumulative density matches the ground truth, the plotted curve will describe a diagonal line with unit slope.
Curves above and below the diagonal {\em underestimate} and {\em overestimate} risk, respectively.
Thus, for the quantitative assessment we calculate the \emph{calibration slope}, which is obtained from the curve described by $1 - \hat{S}_{DKM}(t_i)$ \emph{vs}. $1 - \hat{S}_{KM}(t_i)$.
Since the cumulative density $F(t)$ is unknown for sampling-based approaches, \emph{e.g.}, DATE and SFM, we use a Gaussian Kernel Density Estimator (KDE) \cite{silverman2018density} on samples from the model, $\{t_{ns}\}_{s=1}^{200}$.

Results in Table~\ref{tb:quant} show that in terms of calibration slope fully nonparametric models, namely SFM and DATE, are better calibrated than S-CRPS and DRAFT, both parametrized as log-normally distributed models.
Our approach is the best performing model across all datasets, followed by DATE, S-CRPS, CPH then DRAFT.
We attribute these results to the fact that we directly match the survival function as part of model training.
However, it is surprising that DATE and S-CRPS do not perform nearly as well considering that DATE adversarially matches the time-to-event distribution, thus indirectly matching the cumulative distribution, and S-CRPS that optimizes a proper scoring rule (the integral of Brier score at all possible
thresholds \citep{gneiting2007strictly}) that in principle should produce calibrated predictions.

For the {\sc ehr} data it is not surprising that none of the models are well calibrated because observations in this dataset are not i.i.d. due to patients having multiple encounters.
Since the models and KM-based estimators considered implicitly assume datasets are composed of i.i.d. observations, calibration does not necessarily hold.
This necessitates further investigation, which we leave as interesting future work.
However, to test the hypothesis that the model should be better calibrated in the i.i.d. case, we restricted the {\sc ehr} dataset to the first encounter per patient ($N$=19,064), which results in a better calibrated SFM model (see the Supplementary Material).

{\it Concordance Index}:
C-Index is arguably the most commonly used performance metric in survival analysis.
This metric is useful to assess relative risk because it quantifies ordering rather than temporal accuracy.
Models with high C-Index are good for the purpose of ranking observations into different risk categories, especially in a medical settings.
Since the C-Index is evaluated on point estimates, we summarize time-to-event distributions as medians, \emph{i.e.}, $\hat{t}={\rm median}(\{t_{ns}\}_{s=1}^{200})$, where $t_{ns}$ is a sample from the trained model, $t_{ns}\sim G_{\thetav}(\xv_{n},\epsilonv_s)$, on the test set.

Results in Table~\ref{tb:quant} show that none of the models has a clear advantage over the others, as the C-Index is largely comparable for the remaining four datasets.
Apart from the small and high event rate {\sc support} dataset where S-CRPS and DRAFT (both parametric log-normally distributed models) achieve (statistically) significantly higher C-Index compared to SFM (and CPH).

{\it Coefficient of Variation}:
The Coefficient of Variation (CoV) quantifies the dispersion of a probability distribution.
It is formally defined as $\sigma\mu^{-1}$, where $\sigma$ and $\mu$ are respectively the standard deviation and mean of the distribution being tested.
To summarize the variation of the time-to-event distributions estimated by different models on the test set, we use Mean CoV, which is defined across all time-to-event predictions, \emph{i.e}, $N_{\rm te}^{-1} \sum_{n=1}^{N_{\rm te}} \sigma_n\mu_n^{-1}$, where $N_{\rm te}$ is the size of the test set and $\sigma_i$ and $\mu_i$ are sample standard deviations and means over $\{t_{ns}\}_{s=1}^{200}$.
A model with concentrated time-to-event distributions is one for which mean CoV is as small as possible.
%
\begin{wrapfigure}[21]{R}{0.5\textwidth}
	\begin{minipage}[T]{0.48\textwidth}
		\vspace{-5mm}
		\centering
		\scriptsize
		
		{\sc seer} \\
		\includegraphics[width=0.49\columnwidth]{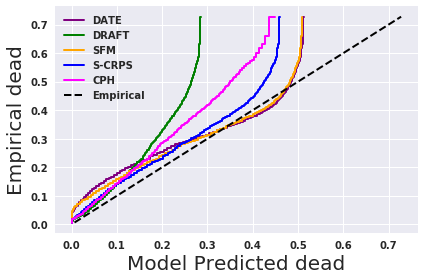}
		\includegraphics[width=0.49\columnwidth]{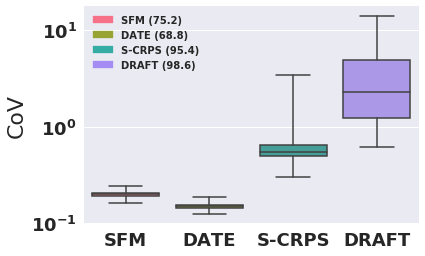}
		
		{\sc sleep} \\
		\includegraphics[width=0.49\columnwidth]{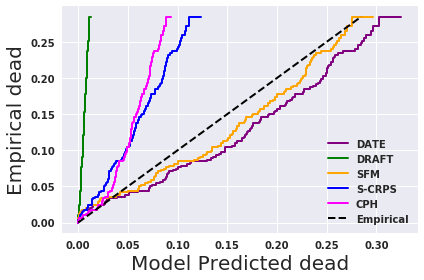}
		\includegraphics[width=0.49\columnwidth]{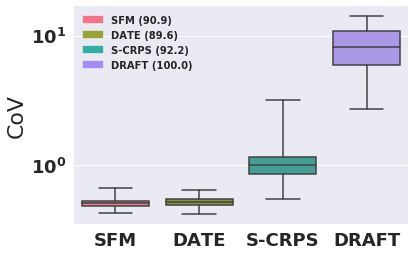}
		\caption{\small Test set calibration and variation visualized for two datasets: {\sc seer} and {\sc sleep} (rows). Left: proportion of events of interest \emph{vs}. predicted events. A perfectly calibrated model will follow the (dashed) diagonal line. Right: coefficient of variation (CoV) distributions. The legend shows the percentage of test set events covered by 95\% intervals from predicted time-to-event distributions.
		}
		\label{fg:calib}
		\vspace{-3mm}
	\end{minipage}
\end{wrapfigure}

Figure~\ref{fg:calib} shows test set CoV distributions.
We see that $i$) DRAFT and S-CRPS have considerably wider variation in CoV thus better 95\% posterior coverage (see legend) compared to SFM and DATE; and $ii$) SFM and DATE are comparable, though DATE is slightly better.
Note that we cannot evaluate CoV or coverage for CPH since in its standard form it only produces point estimates.

Table~\ref{tb:quant} shows that across all datasets DATE, SFM and S-CRPS are on average low-variance models while DRAFT is a considerably higher-variance model.
DATE and SFM are the best-performing in terms predicting concentrated event times given that $\text{mean CoV} < 0.5$.
High-variance time-to-event distributions are not desirable because when prediction uncertainty is large relative to the time range, they cannot be used to inform decision making.
Examples of time-to-event distributions for all models as shown in the Supplementary Material.

\vspace{-3mm}
\paragraph{Qualitative evaluation of calibration}
There are several metrics for measuring the quality of calibration, \emph{e.g.}, calibration slope and Brier score \citep{murphy1973new}.
However, none of these summaries of calibration are as richly informative as visually comparing survival 
functions or cumulative density functions as described above.
In Figure~\ref{fg:calib} we show calibration curves for two different datasets, {\sc seer} and {\sc sleep}, the largest and smallest dataset, respectively.
See the Supplementary Material for figures corresponding to all other datasets including the conditional survival functions as in Figure~\ref{fg:calib_ex}.
From these results (consistent across all datasets) we see that $i$) SFM performs better than the other approaches considered; $ii$) DRAFT is the worst performer; and $iii$) all approaches are poorly calibrated on {\sc seer} data once half of the population has had events.

Under further examination of the {\sc seer} data, we found there is a large subset of the population that gets administratively censored at $t=80$ months (see the Supplementary Material), which explains the generalized sudden divergence of calibration in Figure~\ref{fg:calib}.
This type of \emph{informative} censoring is not random and needs to be modeled appropriately.
However, this extension is beyond the current scope and thus left as future work.
Nonetheless, to test this idea, we truncated the data beyond $t=88$ months and verified that the model is considerably better calibrated (see the Supplementary Material).

\vspace{-2mm}
\section{Conclusions}
\vspace{-1mm}
We have introduced a distribution-based Kaplan-Meier (DKM) estimator for evaluating calibration in time-to-event predictions.
Leveraging this estimator, we introduced SFM, a survival-function-matched neural-network-based model for synthesizing calibrated time-to-event predictions.
Our learning strategy matches the desired survival distribution without the need of an adversarial objective.
Further, we showed that our survival distribution approach is related to earth mover's minimization.
The proposed model outperforms other methods in estimating concentrated and calibrated time-to-event distributions, while remaining competitive in terms of concordance index.
As future work, we plan to extend the proposed approach to calibration in the non-i.i.d. setting, and to account for informative missingness.

\appendix
\section{Coefficient of Variation Results}
\begin{figure}[h!]
	\centering
	\includegraphics[width=0.49\columnwidth]{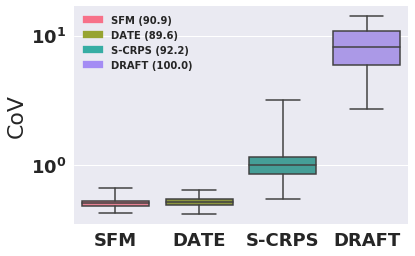}
	\includegraphics[width=0.49\columnwidth]{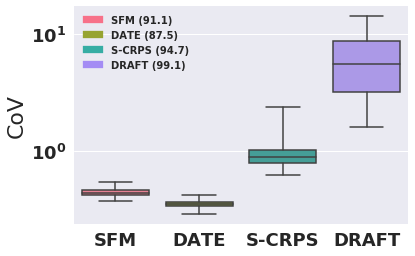}
	\caption{\small  Coefficient of Variation(CoV) distributions for (left) {\sc sleep}  and (right) {\sc flchain} datasets. The legend shows the percentage
		of test set events covered by 95\% intervals from predicted time-to-event distributions.}
	\label{fg:sleep_flchain_cov}
\end{figure}

\begin{figure}[h!]
	\centering
	\includegraphics[width=0.49\columnwidth]{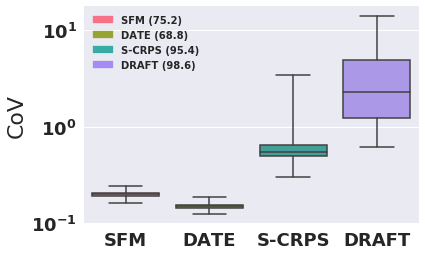}
	\includegraphics[width=0.49\columnwidth]{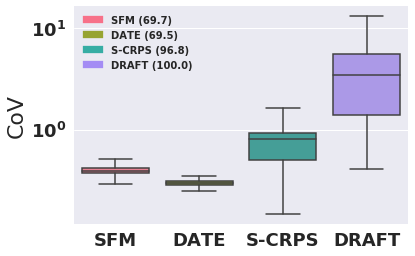}
	\caption{\small  Coefficient of Variation(CoV) distributions for (left) {\sc seer}  and (right) {\sc support} datasets. The legend shows the percentage
		of test set events covered by 95\% intervals from predicted time-to-event distributions.}
	\label{fg:seer_support_cov}
\end{figure}

\begin{figure}[h!]
	\centering
	\includegraphics[width=0.8\columnwidth]{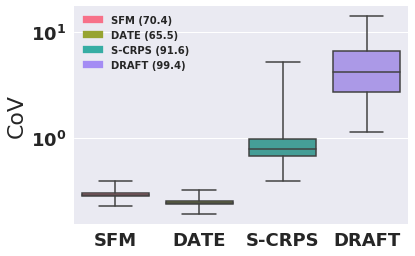}
	\caption{\small  Coefficient of Variation (CoV) distributions for {\sc ehr} dataset. The legend shows the percentage
		of test set events covered by 95\% intervals from predicted time-to-event distributions.}
	\label{fg:ehr_cov}
\end{figure}

See figures \ref{fg:sleep_flchain_cov} , \ref{fg:seer_support_cov} and \ref{fg:ehr_cov}, for  Coefficient of Variation (CoV)  results.

\FloatBarrier
\section{Calibration and Survival Function Results}\label{app:cal}
\begin{figure}[h!]
	\centering
	\includegraphics[width=0.49\columnwidth]{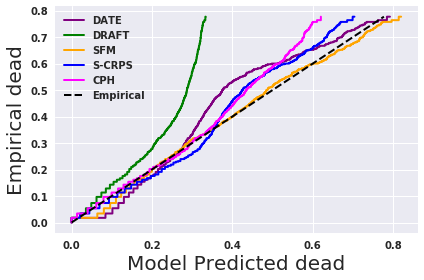}
	\includegraphics[width=0.49\columnwidth]{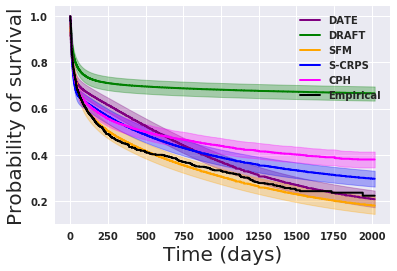}
	\caption{\small  Calibration (left) and Survival function estimates (right) for {\sc support} data. Ground truth (Empirical) is compared to predictions from four models (DATE, DRAFT, SFM (our proposed model), S-CRPS and CPH).}
	\label{fg:support_calib}
\end{figure}

\begin{figure}[h!]
	\centering
	\includegraphics[width=0.49\columnwidth]{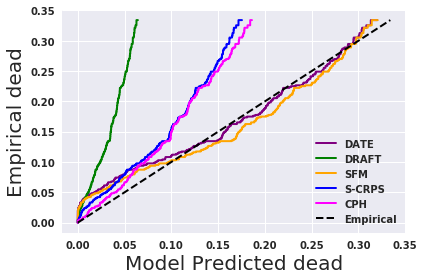}
	\includegraphics[width=0.49\columnwidth]{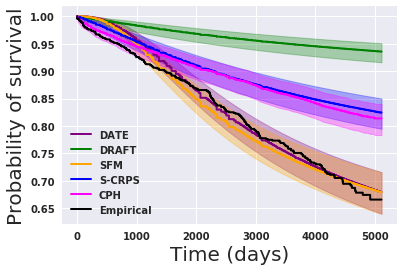}
	\caption{\small Calibration(left) and Survival function estimates (right) for {\sc flchain} data. Ground truth (Empirical) is compared to predictions from four models (DATE, DRAFT, SFM (proposed model), S-CRPS and CPH).}
	\label{fg:flchain_calib}
\end{figure}

\begin{figure}[h!]
	\centering
	\includegraphics[width=0.49\columnwidth]{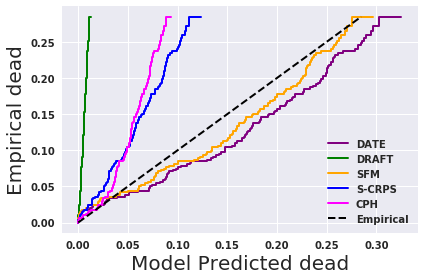}
	\includegraphics[width=0.49\columnwidth]{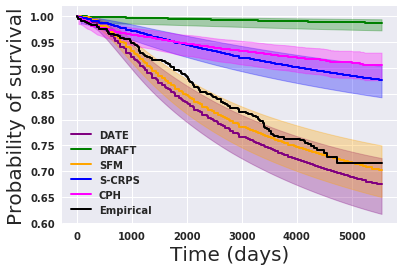}
	\caption{\small Calibration (left) and Survival function estimates (right) for {\sc sleep} data. Ground truth (Empirical) is compared to predictions from four models (DATE, DRAFT, SFM (our proposed model) S-CRPS, and CPH).}
	\label{fg:sleep_calib}
\end{figure}

\begin{figure}[h!]
	\centering
	\includegraphics[width=0.49\columnwidth]{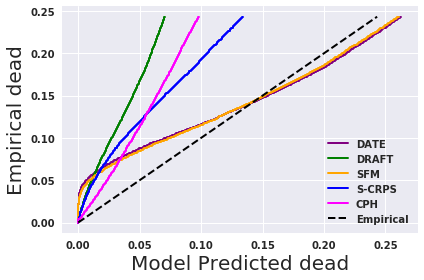}
	\includegraphics[width=0.49\columnwidth]{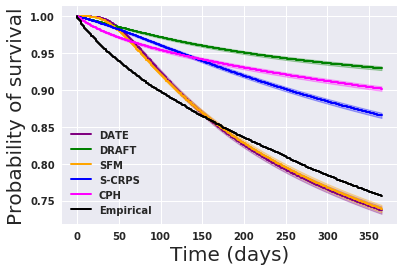}
	\caption{\small Calibration (left) and Survival function estimates (right) for {\sc ehr}  all (non iid) data. Ground truth (Empirical) is compared to predictions from four models (DATE, DRAFT, SFM (our proposed model), S-CRPS, and CPH).}
	\label{fg:ehr_calib}
\end{figure}

\begin{figure}[h!]
	\centering
	\includegraphics[width=0.49\columnwidth]{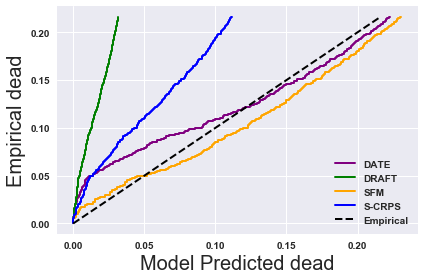}
	\includegraphics[width=0.49\columnwidth]{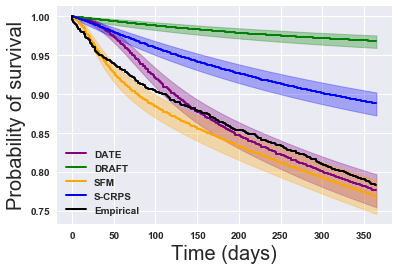}
	\vspace{-3mm}
	\caption{\small Calibration (left) and Survival function estimates (right) for {\sc ehr}  subset  (iid) data. Ground truth (Empirical) is compared to predictions from four models (DATE, DRAFT, SFM (our proposed model) and S-CRPS ).}
	\label{fg:ehr_iid_calib }
\end{figure}

\begin{figure}[h!]
	\centering
	\includegraphics[width=0.49\columnwidth]{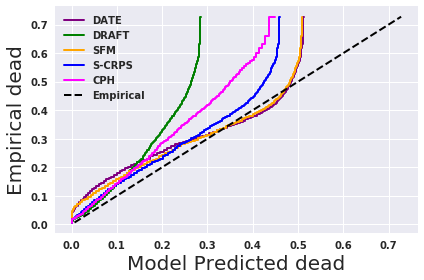}
	\includegraphics[width=0.49\columnwidth]{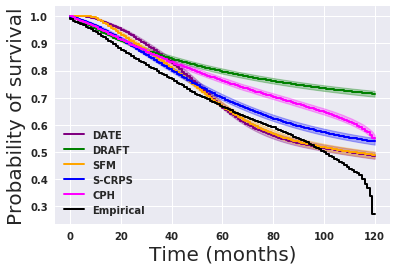}
	\caption{\small Calibration(left) and Survival function estimates (right) for {\sc seer} data. Ground truth (Empirical) is compared to predictions from four models (DATE, DRAFT, SFM (our proposed model), S-CRPS and CPH).}
	\label{fg:seer_calib}
\end{figure}

The model calibration and survival plots for datasets {\sc support}, {\sc flchain}, {\sc sleep}, all {\sc ehr}  (non iid), subset {\sc ehr} (iid) and {\sc seer} are shown in Figures \ref{fg:support_calib}, \ref{fg:flchain_calib}, \ref{fg:sleep_calib},  \ref{fg:ehr_calib}, \ref{fg:ehr_iid_calib } and \ref{fg:seer_calib} respectively.

\subsection{SEER: Informative Censoring}
\begin{figure}[h!]
	\centering
	\includegraphics[width=0.6\columnwidth]{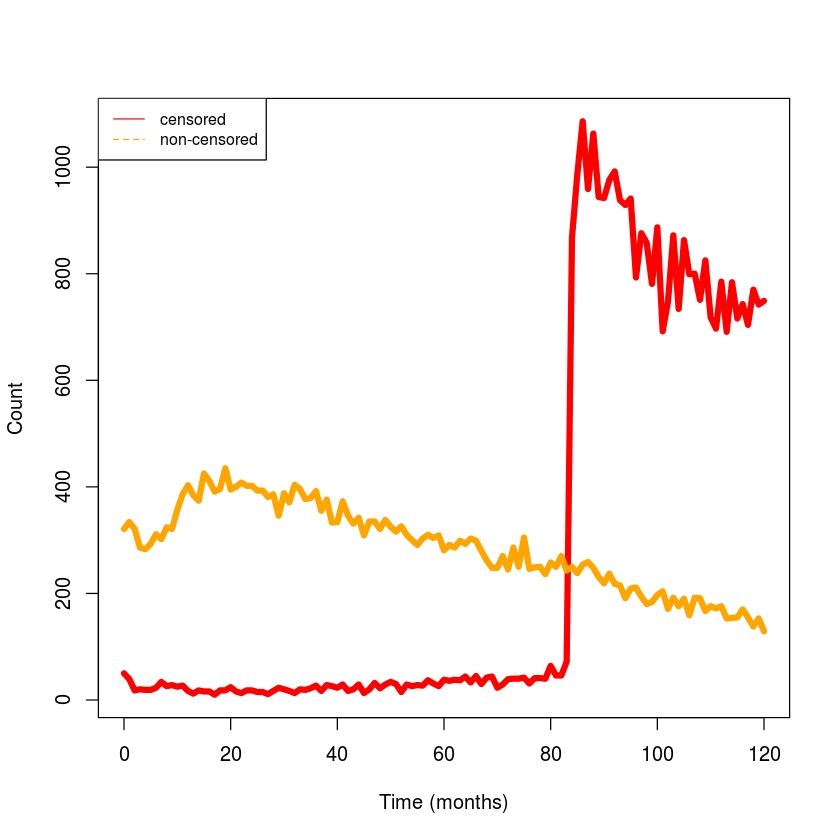}
	\caption{\small Count of censored and non-censored events as a function of time for {\sc seer} data.}
	\label{fg:seer_inf_cens}
\end{figure}

\begin{figure}[h!]
	\centering
	\includegraphics[width=0.49\columnwidth]{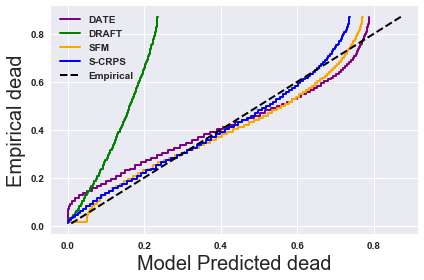}
	\includegraphics[width=0.49\columnwidth]{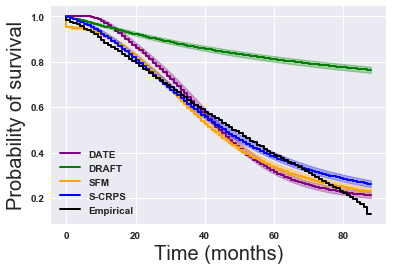}
	\caption{\small \small  Calibration(left) and Survival function estimates (right) for subset {\sc seer} data truncated at 88 months. Ground truth (Empirical) is compared to predictions from four models (DATE, DRAFT, SFM (our proposed model) and S-CRPS).}
	\label{fg:seer_cut_88}
\end{figure}

Figure \ref{fg:seer_inf_cens} shows number of censoring and non-censored events over time.  See Figure \ref{fg:seer_cut_88}, for estimated calibration and survival function results for subset {\sc seer} dataset when truncated at 88 months.

\FloatBarrier
\section{Time-to-Event Distributions }

\begin{figure}[h!]
	\begin{minipage}{\textwidth}
		\centering
		\includegraphics[width=.4\textwidth, height=4.5cm]{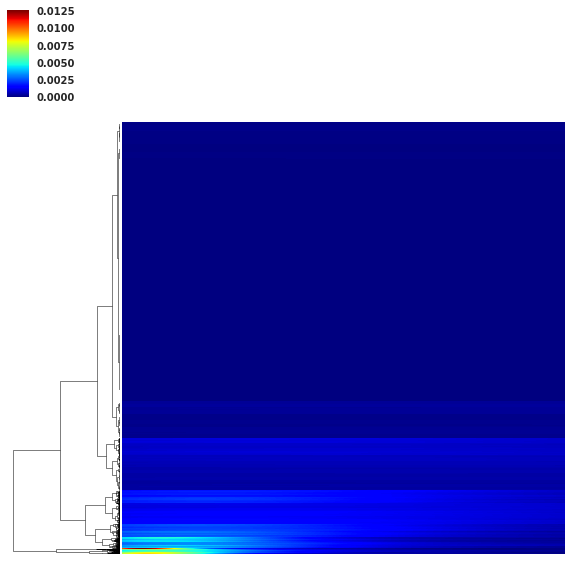}\quad
		\includegraphics[width=.4\textwidth, height=4.5cm]{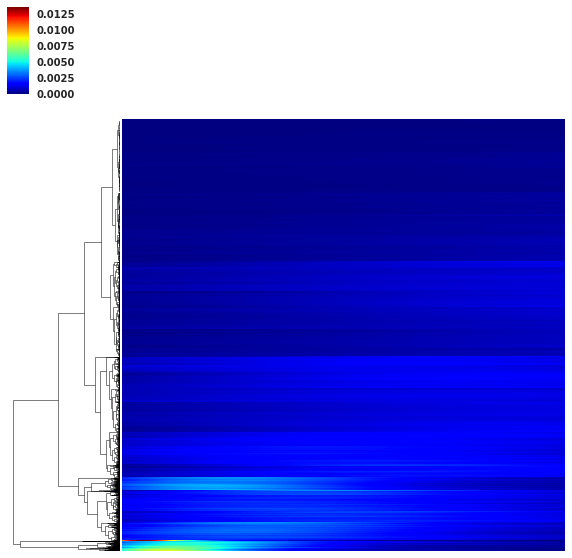}\\
		\includegraphics[width=.4\textwidth, height=4.5cm]{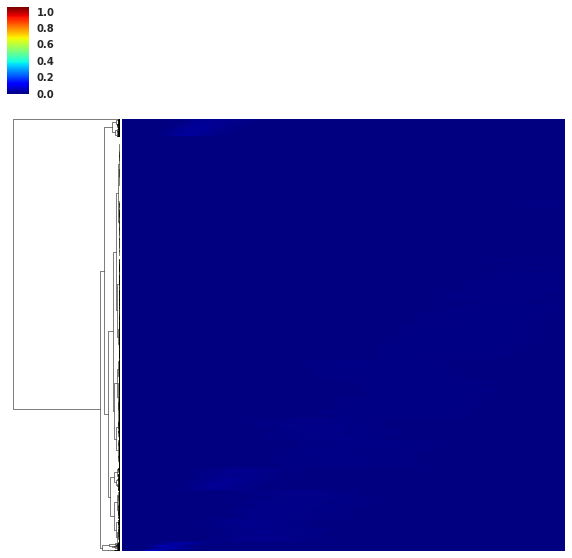}\quad
		\includegraphics[width=.4\textwidth, height=4.5cm]{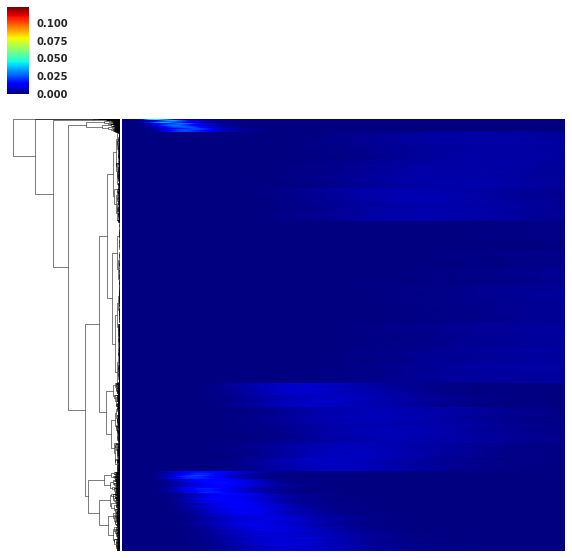}
		\caption{Heatmap of time-to-event distributions on {\sc ehr} data for DRAFT (top-left), S-CRPS (top-right), DATE (bottom-left)  and SFM (bottom-right). The x-axis is the time range $t_{\rm max}$.}
		\label{fig:ehr_hm}
	\end{minipage}
\end{figure}

\begin{figure}[h!]
	\begin{minipage}{\textwidth}
		\centering
		\includegraphics[width=.4\textwidth, height=4.5cm]{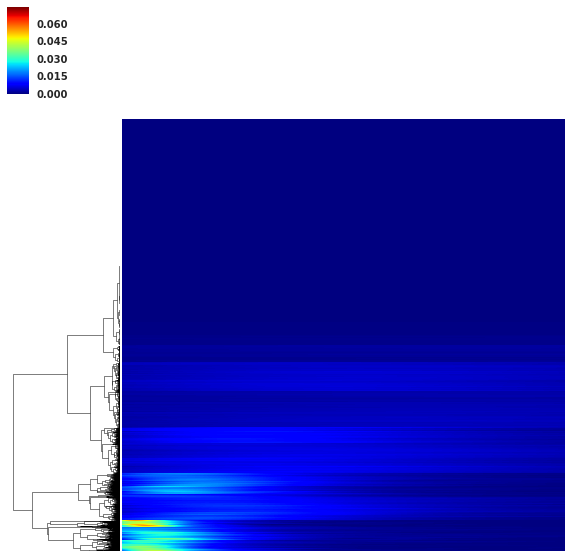}\quad
		\includegraphics[width=.4\textwidth, height=4.5cm]{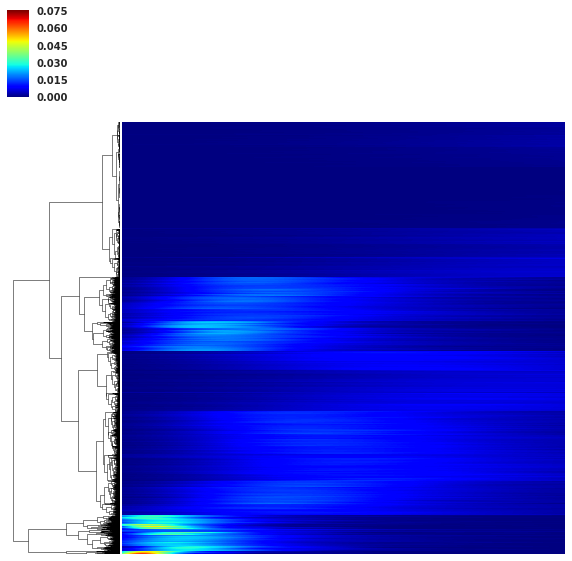}\\
		\includegraphics[width=.4\textwidth, height=4.5cm]{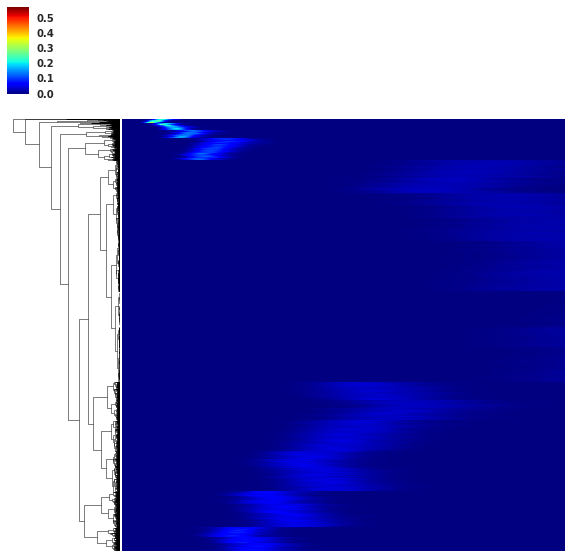}\quad
		\includegraphics[width=.4\textwidth, height=4.5cm]{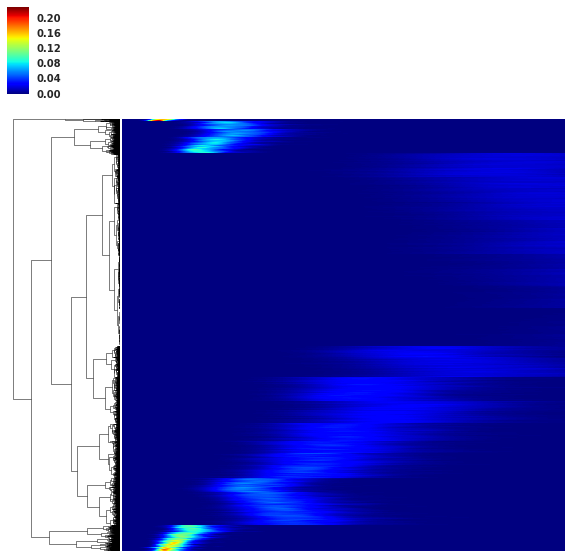}
		\caption{Heatmap of time-to-event distributions on {\sc seer} dataset for DRAFT (top-left), S-CRPS (top-right), DATE (bottom-left)  and SFM (bottom-right). The x-axis is the time range $t_{\rm max}$.}
		\label{fig:seer_hm}
	\end{minipage}\
\end{figure}

\begin{figure}[h!]
	\begin{minipage}{\textwidth}
		\centering
		\includegraphics[width=.4\textwidth, height=4.5cm]{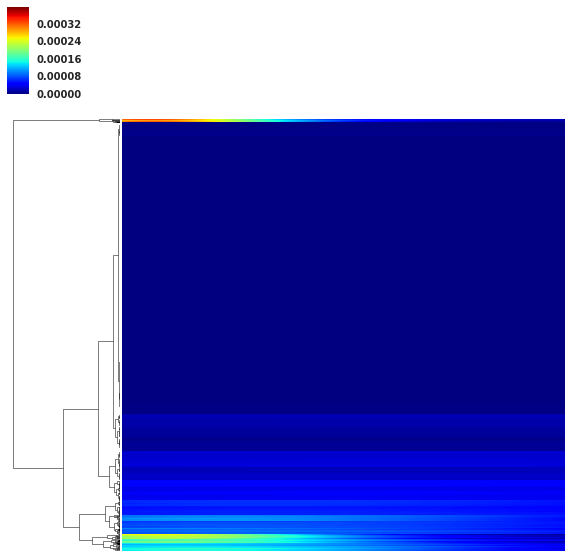}\quad
		\includegraphics[width=.4\textwidth, height=4.5cm]{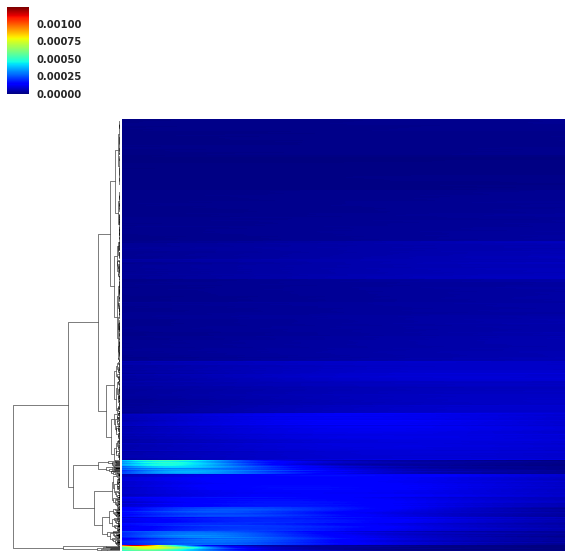}\\
		\includegraphics[width=.4\textwidth, height=4.5cm]{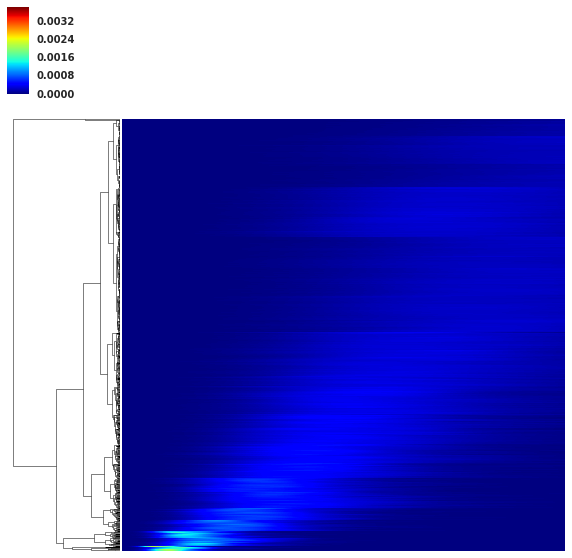}\quad
		\includegraphics[width=.4\textwidth, height=4.5cm]{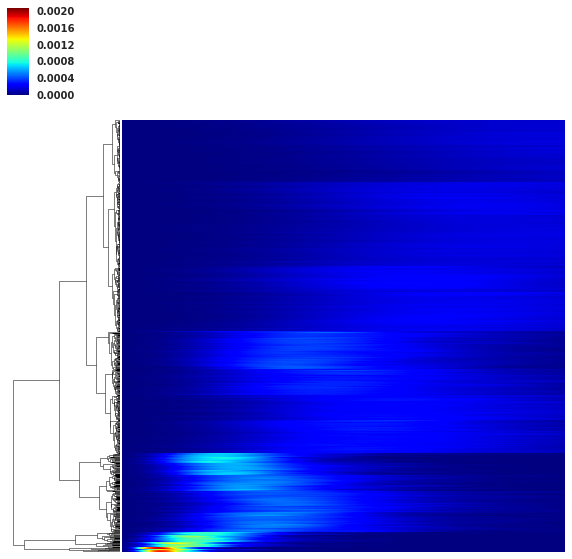}
		\caption{Heatmap of time-to-event distributions on {\sc flchain} data for DRAFT (top-left), S-CRPS (top-right), DATE (bottom-left)  and SFM (bottom-right). The x-axis is the time range $t_{\rm max}$.}
		\label{fig:flchain_hm}
	\end{minipage}
\end{figure}

\begin{figure}[h!]
	\begin{minipage}{\textwidth}
		\centering
		\includegraphics[width=.4\textwidth, height=4.5cm]{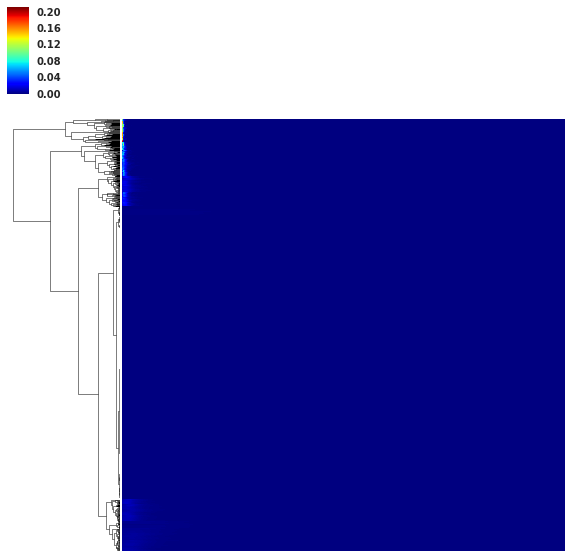}\quad
		\includegraphics[width=.4\textwidth, height=4.5cm]{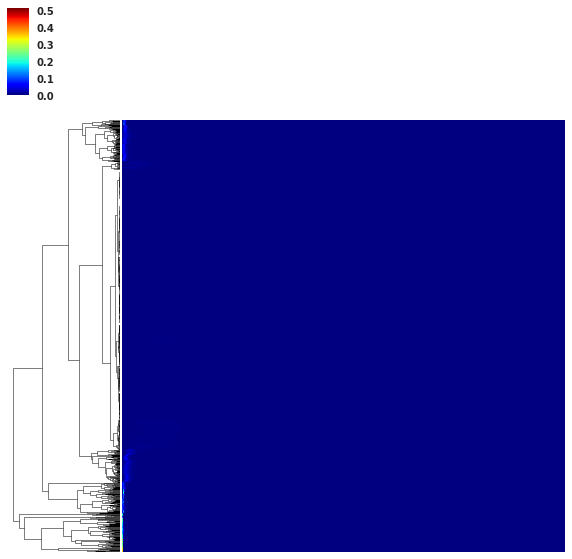}\\
		\includegraphics[width=.4\textwidth, height=4.5cm]{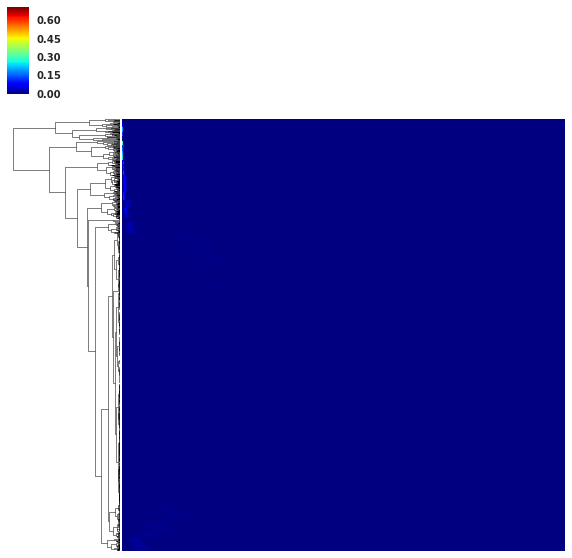}\quad
		\includegraphics[width=.4\textwidth, height=4.5cm]{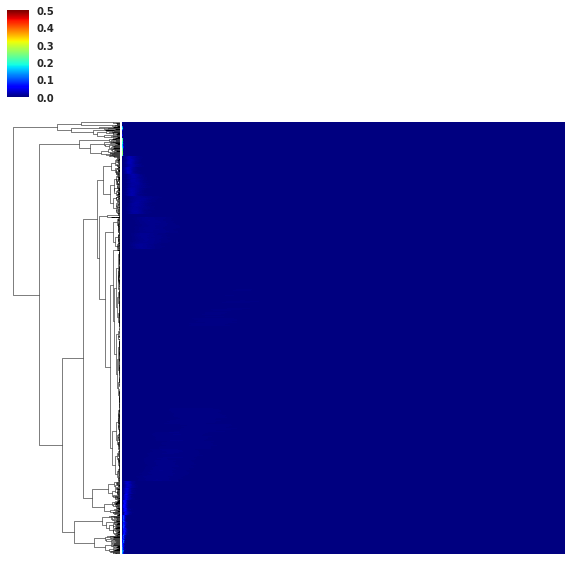}
		\caption{Heatmap of time-to-event distributions on {\sc support} data for DRAFT (top-left), S-CRPS (top-right), DATE (bottom-left)  and SFM (bottom-right). The x-axis is the time range $t_{\rm max}$.}
		\label{fig:support_hm}
	\end{minipage}
\end{figure}

\begin{figure}[h!]
	\begin{minipage}{\textwidth}
		\centering
		\includegraphics[width=.4\textwidth]{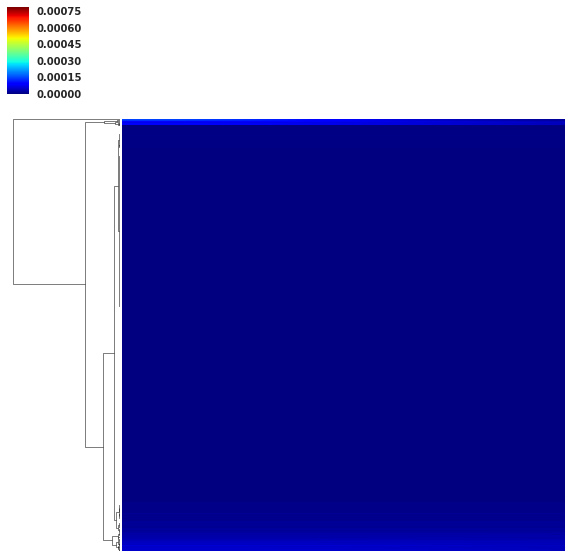}\quad
		\includegraphics[width=.4\textwidth]{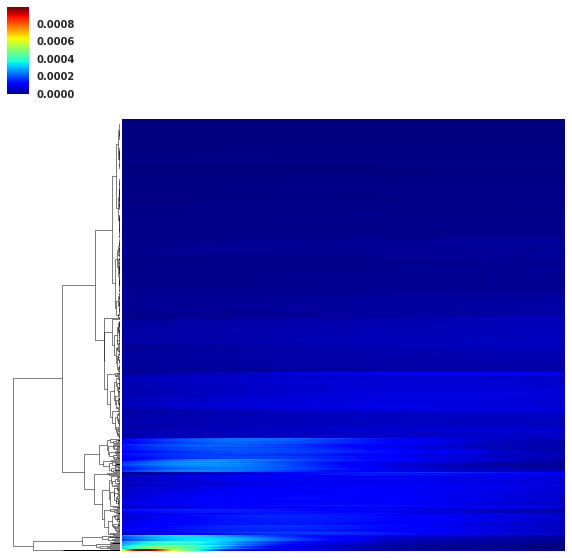}\\
		\includegraphics[width=.4\textwidth]{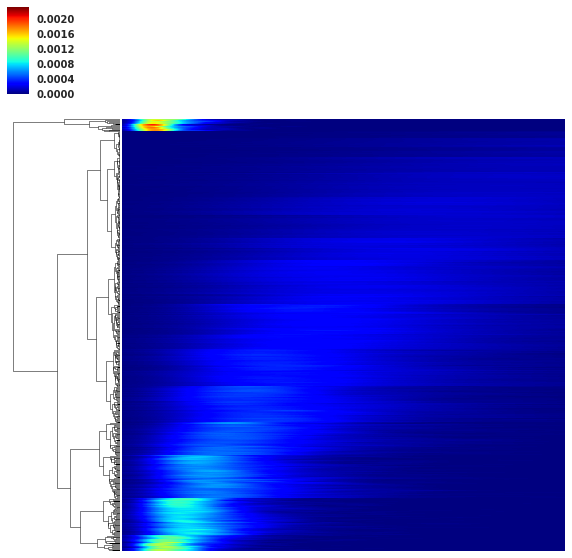}\quad
		\includegraphics[width=.4\textwidth]{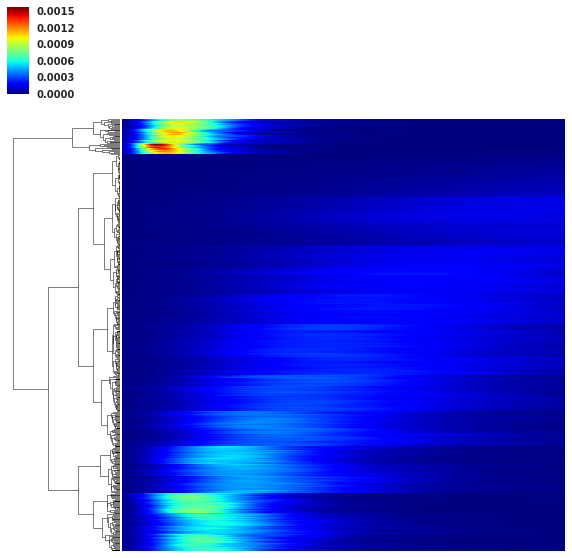}
		\caption{Heatmap of time-to-event distributions on {\sc sleep} data for DRAFT (top-left), S-CRPS (top-right), DATE (bottom-left)  and SFM (bottom-right). The x-axis is the time range $t_{\rm max}$.}
		\label{fig:sleep_hm}
	\end{minipage}
\end{figure}

Figures \ref{fig:ehr_hm}, \ref{fig:seer_hm}, \ref{fig:flchain_hm}, \ref{fig:support_hm} and \ref{fig:sleep_hm}, show the time-to-Event distributions heatmap over the time range $t_{\rm max}$.

\section{Batch Size Sensitivity Analysis}
Table 	\ref{tb:batch}, shows SFM performance metrics across a range of batch sizes.

\begin{table}[h!]
	\centering
	\caption{SFM batch size sensitivity on {\sc flchain} dataset.}
	\vspace{2mm}
	\begin{scriptsize}
		\begin{tabular}{lrrrrr}
			& {\sc 100} & {\sc 250} & {\sc 500} & {\sc 750}  & {\sc 1000}\\
			\toprule
			Calibration slope && & & & \\
			\hdashline
			& 1.0110&0.9766 & 0.9807& 0.9864& \textbf{0.9916} \\
			\toprule
			Mean CoV & & & & &\\
			\hdashline
			&0.4740& \textbf{ 0.4026}&  0.4484&  0.4332& 0.4672\\
			
			\toprule
			C-Index  & & & & &\\
			\hdashline
			& 0.8302& 0.8294& \textbf{0.8318}& 0.8296 & 0.8287 \\
			\bottomrule
			\label{tb:batch}
		\end{tabular}
	\end{scriptsize}
\end{table}

\section{ Experimental Setup}
In all experiments, SFM, DATE, DRAFT  and S-CRPS are specified in terms of two-layer MLPs of $50$ hidden units with Rectified Linear Unit (ReLU) activation functions, batch normalization \cite{ioffe2015batch} and apply dropout of $p=0.2$ on all layers. We set the minibatch size to $M=350$ and use the Adam \cite{kinga2015method} optimizer with the following hyperparameters: learning rate $3 \times 10 ^{-4}$, first moment $0.9$, second moment $0.99$, and epsilon $1 \times 10 ^{-8}$.
We initialize all the network weights according to  \textit{Xavier} \cite{glorot2010understanding}.  SFM  and DATE inject noise in all layers, see \cite{chapfuwa2018adversarial} for more details.
Datasets are split into training, validation and test sets as 80\%, 10\% and 10\% partitions, respectively, stratified by non-censored event proportion. The validation set is used for early stopping and learning model hyperparameters. All models are trained using one NVIDIA P100 GPU with 16GB memory.

\clearpage
{
	{
		\bibliographystyle{abbrv}
		\bibliography{sfm}}
}

\end{document}